\documentclass[11pt,a4paper]{article}
\usepackage{times}
\usepackage{latexsym}
\usepackage{booktabs}

\usepackage{graphicx}
\usepackage{pgfplots}
\usepackage{floatrow}

\usepackage{caption}
\usepackage{subcaption}
\usepackage{floatrow}

\usepackage{url}
\usepackage{makecell}

\usepackage{times}
\usepackage{enumitem}
\usepackage{latexsym}
\usepackage{booktabs}
\usepackage{ltxtable} 
\usepackage{tabularx}
\usepackage{amsmath}
\usepackage{soul}

\usepackage{graphicx}
\usepackage[normalem]{ulem}
\useunder{\uline}{\ul}{}

\usepackage[hyperref]{acl2017}

\aclfinalcopy 

\title{Summary Refinement through Denoising}

\author{
Nikola I. Nikolov, Alessandro Calmanovici, Richard H.R. Hahnloser \\
  Institute of Neuroinformatics, University of Z{\"u}rich and ETH Z{\"u}rich, Switzerland \\
  {\tt \{niniko, dcalma, rich\}@ini.ethz.ch}
}

\date{}

\begin{document}
\maketitle
\begin{abstract}
    We propose a simple method for post-processing the outputs of a text summarization system in order to refine its overall quality. Our approach is to train text-to-text rewriting models to correct information redundancy errors that may arise during summarization. We train on synthetically generated noisy summaries, testing three different types of noise that introduce out-of-context information within each summary. When applied on top of extractive and abstractive summarization baselines, our summary denoising models yield metric improvements while reducing redundancy.\footnote{Code available at \url{https://github.com/ninikolov/summary-denoising}.}
\end{abstract}

\section{Introduction}

Text summarization aims to produce a shorter, informative version of an input text. While extractive summarization only selects important sentences from the input, abstractive summarization generates content without explicitly re-using whole sentences \cite{nenkova2011automatic}. In recent years, a number of successful approaches have been proposed for both extractive \cite{nallapati2017summarunner,refresh} and abstractive \cite{chen2018fast,bottomup} summarization paradigms. Despite these successes, many state-of-the-art systems remain plagued by overly high output redundancy (\citet{see2017get}; see Figure \ref{fig:repetition}), which we set out to reduce.


\begin{figure}[!ht]
    \centering
\includegraphics[width=1\textwidth]{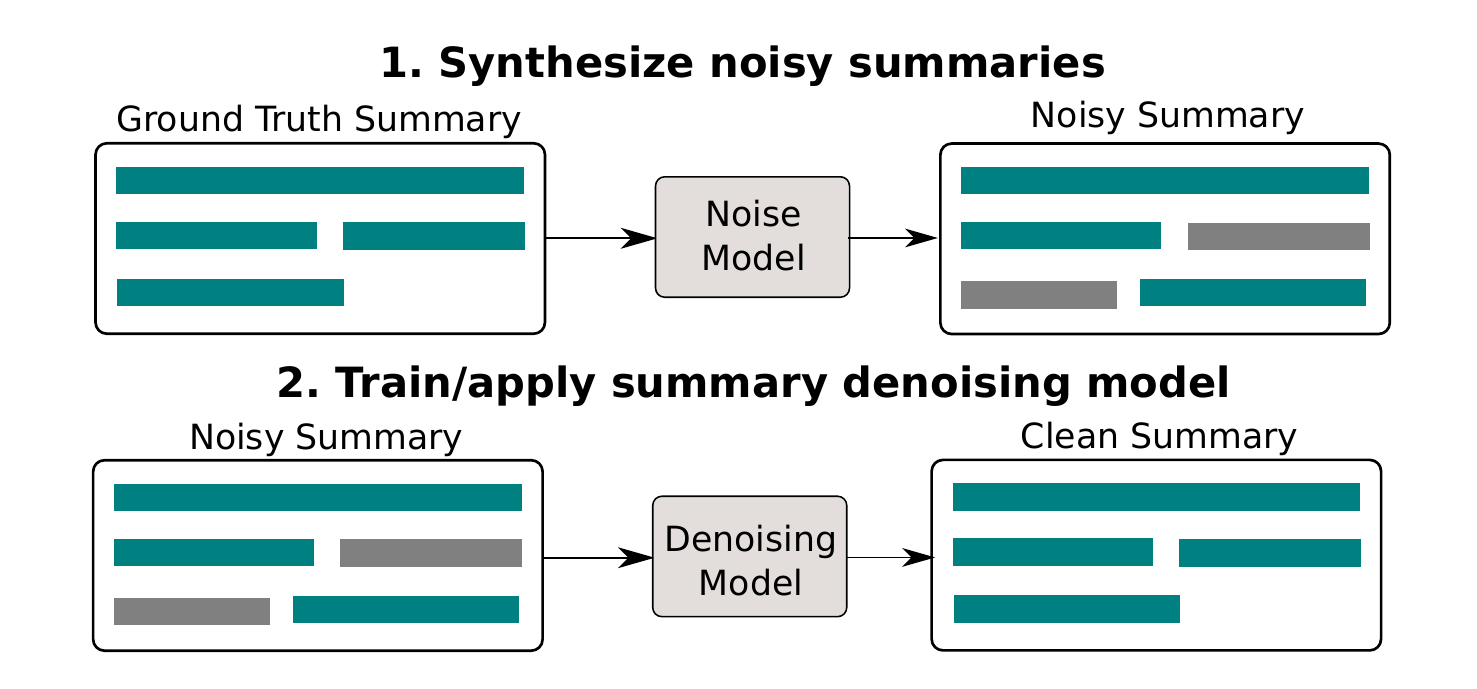}
    \caption{\small{Overview of our approach to summary denoising. We alter ground truth summaries to generate a noisy dataset, on which we train denoising models to restore the original summaries.}}\label{fig:pipeline}
\end{figure}

In this paper, we propose a simple method (Figure \ref{fig:pipeline}, Section \ref{sec:approach}) for post-processing the outputs of a text summarization system in order to improve their overall quality. Our approach is to train dedicated text-to-text rewriting models to correct errors that may arise during summarization, focusing specifically on reducing information redundancy within each individual summary. To achieve this, we synthesize from clean summaries noisy summaries that contain diverse information redundancy errors, such as sentence repetition and out-of-context information (Section \ref{sec:noises}). 

In our experiments (Section \ref{sec:results}), we show that denoising yields metric improvements and reduces redundancy when applied on top of several extractive and abstractive baselines. The generality of our method makes it a useful post-processing step applicable to any summarization system, that standardizes the summaries and improves their overall quality, ensuring fewer redundancies across the text. 

\section{Background}

Post-processing of noisy human or machine-generated text is a topic that has recently been gathering interest. Automatic error correction \cite{rozovskaya2016grammatical,denoising} aims to improve the grammar or spelling of a text. In machine translation, automatic post editing of translated outputs \cite{chatterjee2018findings} is commonly used to further improve the translation quality, standardise the translations, or adapt them to a different domain \cite{isabelle2007domain}.  

In \cite{denoising}, authors synthesize grammatically incorrect sentences from correct ones using backtranslation \cite{sennrich2015improving}, which they use for grammar error correction. They enforce hypothesis variety during decoding by adding noise to beam search. Another work that is close to ours is \cite{compression}, where authors introduce redundancy on the word level in order to build an unsupervised sentence compression system. In this work, we take a similar approach, but instead focus on generating information redundancy errors on the sentence rather than the word level. 

\section{Approach}\label{sec:approach}

Our approach to summary refinement consists of two steps. First, we use a dataset of clean ground truth summaries to \textbf{generate} noisy summaries using several different types of synthetic noise. Second, we train text rewriting models to correct and \textbf{denoise} the noisy summaries, restoring them to their original form. The learned denoising models are then used to post-process and refine the outputs of a summarization system. 

\subsection{Generating noisy summaries}\label{sec:gen-data}

To generate noisy datasets, we rely on an existing parallel dataset of articles and clean ground truth summaries $S=\{s_{0},, ..., s_{j}\}$. We iterate over each of the summaries and perturb them with noise, according to a \textit{sentence noise distribution} $p_{noise} = [p_0, p_1, ..., p_N]$. $p_{noise}$ defines the probability of adding noise to a specific number of sentences within each summary (from $0$ up to a maximum of $N$ noisy sentences), with $\sum p_{noise} = 1$.  

For all experiments in this work, we use $p_{noise} = [0.15, 0.85]$ in order to ensure consistency, meaning that \textasciitilde15\% of our noisy summaries contain no noisy sentences, while \textasciitilde85\% contain one noisy sentence. Initial experiments showed that distributions which enforce larger or smaller amounts of noise lead to stronger or weaker denoising effects. Our choice of noise distribution showed good results on the majority of systems that we tested; we leave a more rigorous investigation of the choice of distribution to future work.

In addition to adding noise, we generate 3 noisy summaries for each clean summary by picking multiple random sentences to noise. This step increases the dataset size while introducing variety. 

\subsection{Types of noise}\label{sec:noises}

We experiment with three simple types of noise, all of which introduce \textit{information redundancy} into a summary. Our aim is to train denoising models that minimize repetitive or peripheral information within summaries. 

\paragraph{Repeat} picks random sentences from the summary and repeats them at the end.  Repetition of phrases or even whole sentences is a problem commonly observed in text generation with RNNs \cite{see2017get}, which motivates efforts to detect and minimize repetitions. 

\paragraph{Replace} picks random sentences from the summary, and replaces them with the closest sentence from the article. This type of noise helps the model to learn to \textit{refine} sentences from the generated summaries, paraphrasing sentences when they are too long or contain redundant information. 

\paragraph{Extra} picks random sentences from the article, paraphrases them, and inserts them into the summary, preserving the order of the sentences as they appear in the article. With this type of noise, a model learns to delete sentences which are out of context or contain redundant information. To paraphrase the sentences, we use the sentence paraphrasing model from \cite{chen2018fast}, trained on matching sentence pairs from the CNN/Daily Mail dataset. 

\paragraph{Mixture} mixes all the above noise types uniformly into a single dataset, keeping the same dataset size as for the individual noise types. With mixture, we explore whether the benefits of each noise type can be combined into a single model. 

\begin{figure*}[!ht]
\centering
 \hspace*{-1cm}
\begin{subfigure}[b]{0.55\textwidth}
  \centering
  \includegraphics[width=\textwidth]{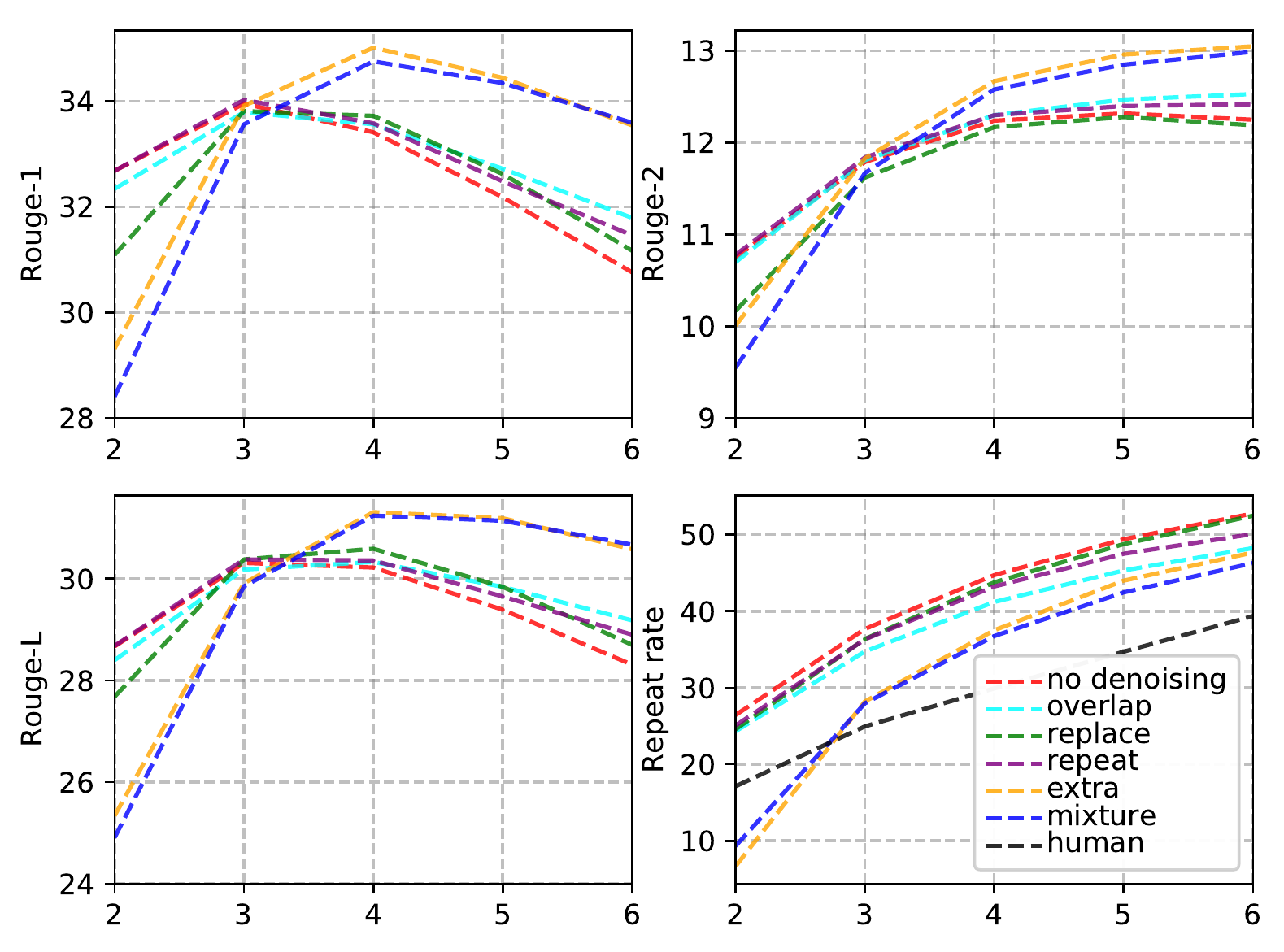}
  \caption{Denoising the LexRank system.}
  \label{fig:lexrank}
\end{subfigure}%
\hspace*{-0.23cm}
\begin{subfigure}[b]{0.55\textwidth}
  \centering
  \includegraphics[width=\textwidth]{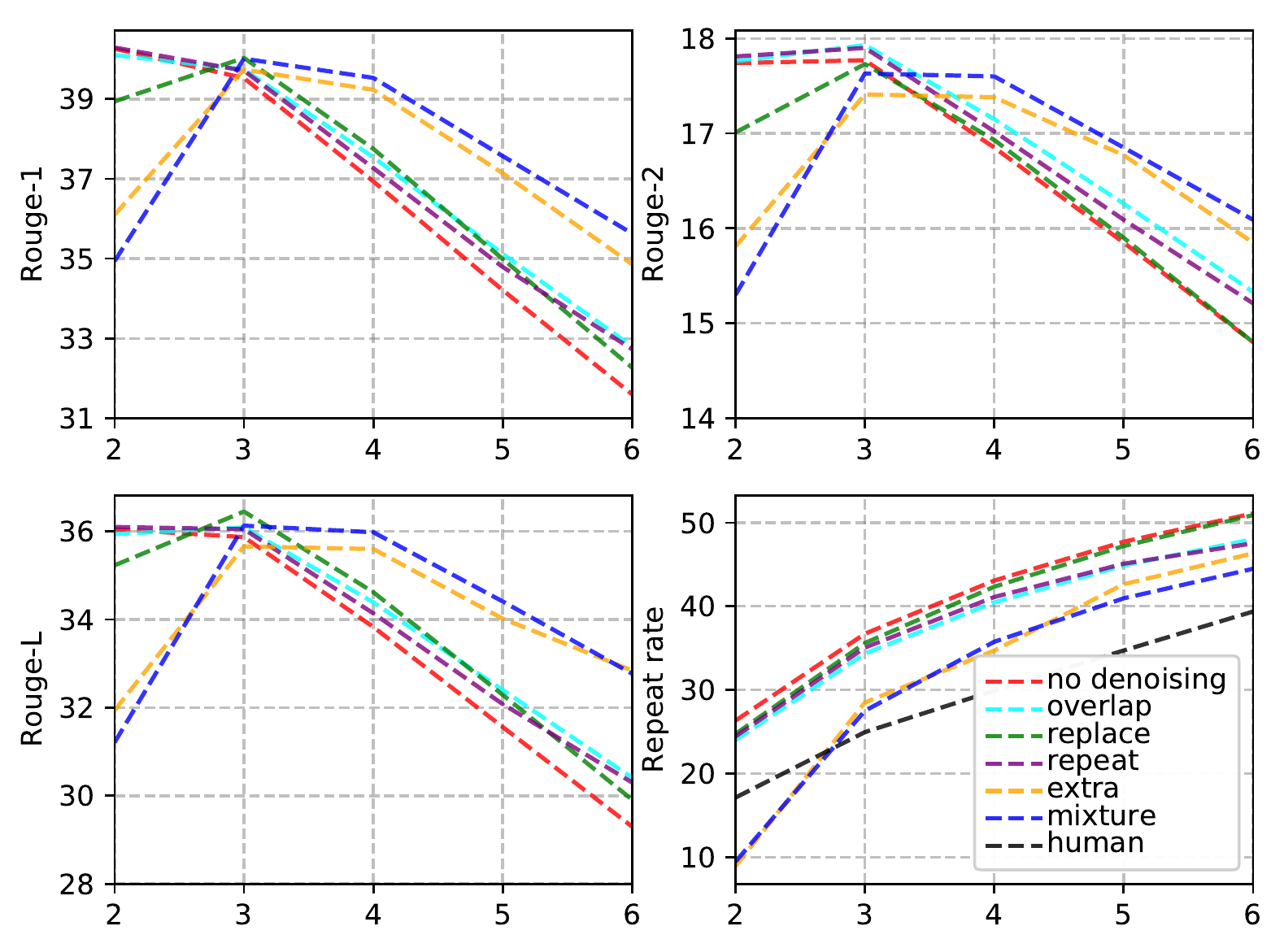}
  \caption{Denoising the RNN-Ext system.}
  \label{fig:extrnn}
\end{subfigure}
\caption{\small{Metric results (Rouge-1/2/L and Repeat rate) on denoising extractive summarization systems. The x-axis in all plots is the number of extracted sentences. \texttt{human} is the result of the ground truth summaries (only for the Repeat rate).}}
\label{fig:extractive}
\end{figure*}

\section{Experimental set-up}\label{sec:setup}

\paragraph{Dataset} 
We use the CNN/Daily Mail dataset\footnote{\url{https://github.com/abisee/cnn-dailymail}} \cite{hermann2015teaching} of news articles and summaries in the form of bullet points, and follow the preprocessing pipeline from \cite{chen2018fast}. We use the standard split of the dataset, consisting of 287k news-summary pairs for training and 13k pairs for validation. We follow Section \ref{sec:gen-data} to generate noisy versions of the datasets to be used during training. During testing, instead of clean summaries that contain noisy sentences, we input summaries produced by existing extractive or abstractive summarization systems. 

\paragraph{Denoising models}

For all of our denoising experiments, we use a standard bidirectional LSTM encoder-decoder model \cite{sutskever2014sequence} with 1000 hidden units and an attention mechanism \cite{bahdanau2014neural}, and train on the subword-level \cite{sennrich2015neural}, capping the vocabulary size to 50k tokens for all experiments\footnote{We use the \texttt{fairseq} library \url{https://github.com/pytorch/fairseq}}. We train all models until convergence using the Adam optimizer \cite{kingma2014adam}. 

In addition to our neural denoising models, we implement a simple denoising baseline, \texttt{overlap}, based on unigram overlap between sentences in a summary. \texttt{overlap} deletes sentences which overlap more than $80\%$\footnote{We empirically found that this threshold is sufficiently high to prevent unnecessary deletion and sufficiently low to detect near-identical sentences.} with any other sentence in the summary and can therefore be considered as redundant. 

\paragraph{Evaluation} We report the ROUGE-1/2/L metrics \cite{lin2004rouge}. We also report the \textit{Repeat} rate \cite{science-summ} $rep(\mathbf{s})=\frac{\sum_i o(\overline{\mathbf s_i}, s_{i})}{|\mathbf{s}|}$  which is the average unigram overlap $o$ of each sentence $s_{i}$ in a text with the remainder of the text (where $\overline{\mathbf s_i}$ denotes the complement of sentence $s_{i}$). Since the repeat rate measures the overlapping information across all sentences in a summary, lower values signify that a summary contains many unique sentences, while higher values indicate potential information repetition or redundancy within a summary. 

\section{Results}\label{sec:results}

\subsection{Extractive summarization}

We experiment with denoising two extractive systems: LexRank \cite{erkan2004lexrank} is an unsupervised graph-based approach which measures the centrality of each sentence with respect to the other sentences in the document. RNN-Ext is a more recent supervised LSTM sentence extractor module from \cite{chen2018fast}, trained on the CNN/Daily Mail dataset. It extracts sentences from the article sequentially. Both extractive systems require the number of sentences to be extracted to be given as a hyperparameter, in our experiments we test with summary lengths ranging from $2$ to $6$ sentences\footnote{The average sentence count of a summary in the CNN/Daily Mail dataset is $3.88$.}. 

\begin{table*}
\RawFloats
		\centering
		\small 
\begin{tabular}{c|c|c|c|c|c|c|c}
\Xhline{3\arrayrulewidth}
\textbf{System} & \textbf{Denoising approach} & \textbf{ROUGE-1} & \textbf{ROUGE-2} & \textbf{ROUGE-L} & \textbf{Repeat} & \textbf{\#Sent} & \textbf{\#Tok} 
\\ \Xhline{3\arrayrulewidth}
Human & - & - & - & - & 28.86 & 3.88 & 61.21
\\ \Xhline{2\arrayrulewidth}
Article & - & 14.95 & 8.54 & 14.41 & 70.5 & 26.9 & 804 \\ \hline
Article & Mixture & 30.47 & 13.97 & 28.24 & 53.43 & 10.67 & 304.7 
\\ \Xhline{2\arrayrulewidth}
RNN & - & 35.61 & 15.04 & 32.7 & 51.9 & 2.93 & 58.46 \\ \hline
RNN & Overlap & 36.41 & 15.92 & 33.73 & 26.84 & 2.39 & 47.31 \\ \hline
RNN & Repeat & \textbf{36.5} & \textbf{15.94} & \textbf{33.79} & 27.65 & 2.41 & 48.34 \\ \hline
RNN & Replace & 35.2 & 14.86 & 32.4 & 51.51 & 2.98 & 57.0 \\ \hline
RNN & Extra & 33.95 & 14.58 & 31.2 & 37.19 & 2.21 & 42.82 \\ \hline
RNN & Mixture & 35.08 & 15.3 & 32.44 & 27.27 & 2.2 & 42.14 \\ \Xhline{2\arrayrulewidth}
RNN-RL & - & \textbf{40.88} & \textbf{17.8} & \textbf{38.54} & 39.29 & 4.93 & 72.82 \\ \hline
RNN-RL & Overlap & 40.76 & 17.69 & 38.43 & 37.71 & 4.83 & 71.02 \\ \hline
RNN-RL & Repeat & 40.84 & 17.76 & 38.49 & 38.78 & 4.86 & 71.69 \\ \hline
RNN-RL & Replace & 40.78 & 17.72 & 38.46 & 39.24 & 4.93 & 72.2 \\ \hline
RNN-RL & Extra & 39.12 & 16.7 & 36.76 & 34.04 & 3.84 & 55.43 \\ \hline
RNN-RL & Mixture & 40.11 & 17.33 & 37.76 & 35.45 & 4.18 & 61.15 \\ \Xhline{3\arrayrulewidth} 
\end{tabular}
\caption{\small{Results on denoising abstractive summarization. \textbf{Repeat} is the Repeat rate, while \textbf{\#Sent} and \textbf{\#Tok} are the average numbers of sentences or tokens in the summaries. Best \textbf{ROUGE} results for each model are in bold. \textit{Human} is the result of the ground truth summaries, while \textit{Article} uses the original article as the summary.}}
\label{tab:abstractive}
\end{table*}

\begin{figure*}[!ht]
\centering
\begin{subfigure}[b]{0.5\textwidth}
  \centering
  \includegraphics[width=\textwidth]{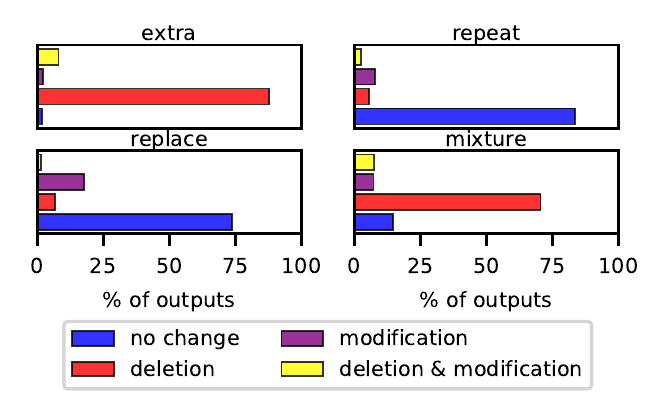}
  \caption{RNN-Ext extractive system, extracting 5 sentences.}
  \label{fig:operations-rnn-ext}
\end{subfigure}%
\begin{subfigure}[b]{0.5\textwidth}
  \centering
  \includegraphics[width=\textwidth]{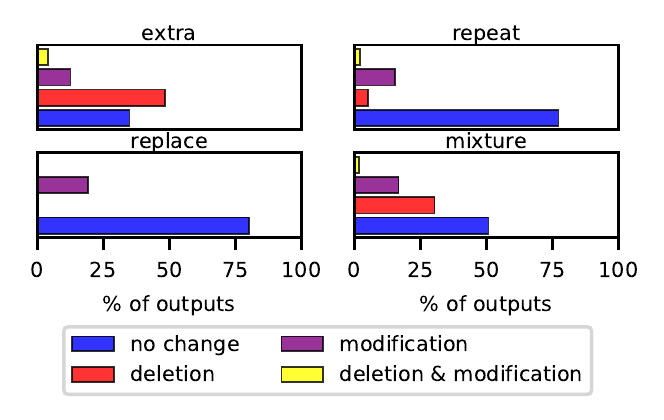}
  \caption{RNN abstractive system.}
  \label{fig:operations-rnn}
\end{subfigure}
\caption{\small{Types of denoising operations applied to an extractive (left) and an abstractive (right) system (averaged over our test set).}}
\label{fig:operations}
\end{figure*}

The results on extractive summarization are in Figure \ref{fig:lexrank} for \textit{LexRank} and Figure \ref{fig:extrnn} for \textit{RNN-Ext}, where we plot the metric scores for varying numbers of extracted sentences for each of the two systems. For both LexRank and RNN-Ext, we observe ROUGE improvements after denoising over the baseline systems without denoising. The \texttt{repeat} and \texttt{replace} methods yielded more modest improvements of 0.5-1 ROUGE-L points, performing comparably to the simple \texttt{overlap} baseline. The most effective noise types are \texttt{extra} and \texttt{mixture}, yielding improvements of up to 2 ROUGE-L points for LexRank and up to 3.5 ROUGE-L points for RNN-Ext. The superior performance to \texttt{overlap} indicates that the additional denoising operations learned by our models (see Figure \ref{fig:operations-rnn-ext}) are beneficial and can lead to more polished summaries that also may contain abstractive elements. 

\begin{figure}
    \centering
    \includegraphics[width=0.9\textwidth]{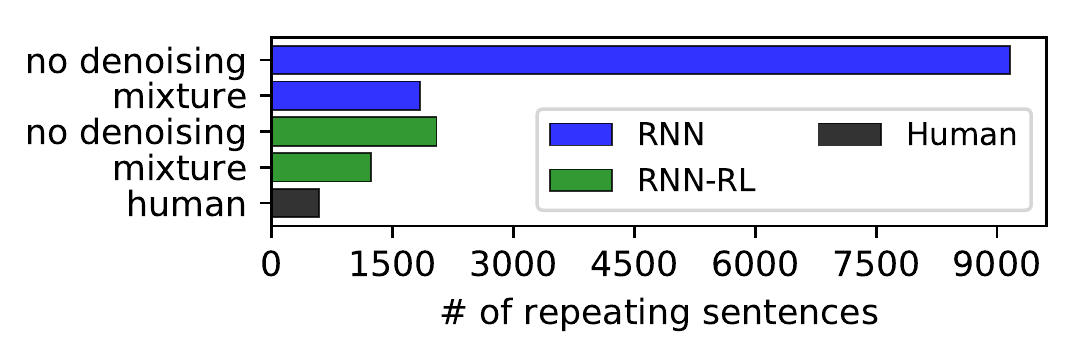}
    \caption{\small{Number of sentence repetitions before and after denoising.}}
    \label{fig:repetition}
\end{figure}

The gains from denoising are greater for longer summaries of more than two sentences. Long summaries are more likely to be affected by redundancy. For shorter summaries, denoising might lead to deletion of important information, thus denoising needs to be applied more carefully in such cases. Furthermore, for all sentence lengths and noise types, we observe a reduction in the Repeat rate after denoising, demonstrating that our approach is effective at reducing redundancy. 

In Table \ref{tab:abstractive}, we additionally include the result from using the whole articles (\textit{Article}) as input to our \texttt{mixture} model. Denoising is effective in this case, indicating that our approach may be promising for developing abstractive summarization systems that are fully unsupervised, similar to recent work in unsupervised sentence compression \cite{compression}. 

\subsection{Abstractive summarization}

For abstractive summarization, we test two systems. The first is a standard LSTM encoder-decoder model with an attention mechanism (\textit{RNN}), identical to our denoising network from Section \ref{sec:setup}. The second, \textit{RNN-RL}, is a state-of-the-art abstractive system proposed in \cite{chen2018fast} that combines extractive and abstractive summarization using reinforcement learning. We train \textit{RNN} ourselves, while for \textit{RNN-RL}, we use the outputs provided by the authors. 

Our metric results from denoising abstractive summarization are in Table \ref{tab:abstractive}. In Figure \ref{fig:repetition}, we also compute the approximate number of sentence repetitions on the test set, by calculating the number of sentences that overlap significantly ($>80\%$) with at least one other sentence in the summary. 

For the \textit{RNN} model, the \texttt{repeat} noise helps to remove repetition, halving our repetition metric, while boosting the ROUGE scores. This result is similar to our much simpler \texttt{overlap} baseline based on sentence deletion. The other noise types help to reduce redundancy, bringing the \textit{Repeat} rate closer to that of \textit{Human} summaries. This, however, comes at the cost of a decrease in ROUGE. For \textit{RNN-RL}, while denoising helps to reduce repetition, none of our noise types managed to yield ROUGE improvements. One reason for this may be that this model already comes with a built-in mechanism for reducing redundancy which relies on sentence reranking \cite{chen2018fast}. However, as shown in Figure \ref{fig:repetition} (and in our example in Table \ref{table:example}), this model still generates many more sentence repetitions than found in human summaries. 
In overall, our approach is effective at reducing redundant information in abstractive summaries, however this comes with a potential loss of information, which can lead to a reduction in ROUGE. Thus, our denoising methods are currently better suited for extractive than for absctractive summarization. Our work therefore calls for the development of novel types of synthetic noise that target abstractive summarization.

\subsection{Analysis of model outputs}

In Figure \ref{fig:operations}, we quantify the types of operations (deletion or modification of one or more sentences, or no change) our denosing models performed on the summaries produced by the extractive \textit{RNN-Ext} (Figure \ref{fig:operations-rnn-ext}) and abstractive \textit{RNN} system (Figure \ref{fig:operations-rnn}). The \texttt{replace} and \texttt{repeat} noises are the most conservative, leaving over 75\% of the summaries unchanged. \texttt{extra} is the most prone to delete sentences, while \texttt{repeat} and \texttt{replace} are most prone to modify sentences. We see a similar pattern for both extractive and abstractive summarization, with an increase of deletion for longer summaries produced by the extractive system. This indicates that our approach flexibly learns to switch between operations depending on the properties of the noisy input summary. 

In Table \ref{table:example} we show example outputs from denoising extractive and abstractive summaries produced for a sports article from our test set. All baseline summarization systems produced outputs that contain redundancy: for example, the first three sentences generated by the \textit{RNN} system, and the 3rd and 4th sentences produced by the \textit{RNN-RL} system are almost identical. To denoise the summaries, our models used diverse operations such as deletion of one or two sentences (e.g. \textit{RNN} system, \textit{Repeat} noise), rewriting (e.g. \textit{RNN-RL} system, \textit{Replace} noise, where \textit{"dinorah santana , the player ’s agent , said her client had rejected the offer of a three-year contract extension"} is paraphrased to \textit{"the player ’s agent said she had rejected the offer of a three-year contract"}), or even a combination of deletion and rewriting (e.g. \textit{RNN-RL} system, \textit{Repeat} noise). 

\section{Conclusion}

We proposed a general framework for improving the outputs of a text summarization system based on denoising. Our approach is independent of the type of the system, and is applicable to both abstractive and extractive summarization paradigms. It could be useful as a post-processing step in a text summarization pipeline, ensuring that the summaries meet specific standards related to length or quality. 

Our approach is effective at reducing information repetition present in existing summarization systems, and can even lead to ROUGE improvements, especially for extractive summarization. Denoising abstractive summarization proved to be more challenging, and our simple noise types did not yield significant ROUGE improvements for a state-of-the-art system. Our focus in future work, will, therefore, be to estimate better models of the noise present in abstractive summarization, to reduce information redundancy without a loss in quality, as well as to target other aspects such as the grammaticality or cohesion of the summary.


\newcommand\setItemnumber[1]{\setcounter{enumi}{\numexpr#1-1\relax}}
{\fontsize{6}{6}\selectfont
\renewcommand{\arraystretch}{1}
\begin{table*}
\centering
\caption{\small{Examples for denoising extractive and abstractive summarization. \textit{Same} indicates a summary sentence has been unchanged, while \textit{Deleted} indicates sentence deletion. In brackets, \textit{R-1} denotes the \textit{Rouge-1} score, while \textit{Rep} denotes the \textit{Repeat} rate.}}
\label{table:example}
\begin{tabularx}{1.0\textwidth}{|X|X|X|}
\hline
\multicolumn{3}{|p{0.95\textwidth}|}{\centering\small{\textbf{Ground truth} \textit{(Rep=38.38)}\textbf{:} \begin{enumerate}[topsep=0pt,itemsep=-0.5pt,after=\vspace{-0.4cm},leftmargin=.24\linewidth]
  \setlength\itemsep{-0.5em}
  \item dani alves has spent seven seasons with the catalan giants
  \item alves has four spanish titles to his name with barcelona
  \item the brazil defender has also won the champions league twice with barca
\end{enumerate} }} 
\\ 
\hline
\small{\textbf{RNN-Ext-4}} & \small{\textbf{RNN}} & \small{\textbf{RNN-RL}} \\ \hline
\small{ 
\textit{No denoising (R-1=33.6,Rep=45):}

\begin{enumerate}[topsep=-2pt,itemsep=-0.5pt, after=\vspace{-0.4cm},wide, labelwidth=!, labelindent=0pt]
\setlength\itemsep{-0.5em}
  \item dani alves looks set to leave barcelona this summer after his representative confirmed the brazilian right-back had rejected the club 's final contract offer
  \item alves has enjoyed seven successful years at barcelona , winning four spanish titles and the champions league twice
  \item but the 31-year-old has been unable to agree a new deal with the catalan club and will leave the nou camp this summer
  \item dinorah santana , the player 's agent and ex-wife , said at a press conference on thursday that her client had rejected the offer of a three-year contract extension , which was dependent on the player taking part in 60 per cent of matches for the club
\end{enumerate}} & \small{\textit{No denoising (R-1=34,Rep=79.6):}
\begin{enumerate}[topsep=-2pt,itemsep=-0.5pt, after=\vspace{-0.4cm},wide, labelwidth=!, labelindent=0pt]
\setlength\itemsep{-0.5em}
  \item dani alves has been unable to agree a new deal with catalan club
  \item the brazilian has been unable to agree a new deal with catalan club  
  \item alves has been unable to agree a new deal with catalan club
  \item alves has been linked with a number of clubs including manchester united and manchester city
\end{enumerate}} & \small{\textit{No denoising (R-1=31,Rep=51.6):}
\begin{enumerate}[topsep=-2pt,itemsep=-0.5pt, after=\vspace{-0.4cm},wide, labelwidth=!, labelindent=0pt]
\setlength\itemsep{-0.5em}
  \item dani alves looks set to leave barcelona this summer
  \item alves has enjoyed seven successful years at barcelona  
  \item alves has been unable to agree a deal with the catalan club
  \item the 31-year-old has been unable to agree a new deal
  \item dinorah santana , the player 's agent , said her client had rejected the offer of a three-year contract extension
\end{enumerate}} \\ \hline
\small{\textit{Replace (R-1=36.6,Rep=46.6):}
\begin{enumerate}[topsep=-2pt,itemsep=-0.5pt, after=\vspace{-0.4cm},wide, labelwidth=!, labelindent=0pt]
\setlength\itemsep{-0.5em}
  \item Same
  \item Same
  \item Same
  \item the player 's agent and ex-wife said at a press conference on thursday that her client had rejected the offer of a three-year contract extension
\end{enumerate}} & \small{\textit{Replace (R-1=34, Rep=79.6):}
\begin{enumerate}[topsep=-2pt,itemsep=-0.5pt, after=\vspace{-0.4cm},wide, labelwidth=!, labelindent=0pt]
\setlength\itemsep{-0.5em}
  \item Same
  \item Same
  \item Same
  \item Same
\end{enumerate}} & \small{\textit{Replace (R-1=31, Rep=52.6):}
\begin{enumerate}[topsep=-2pt,itemsep=-0.5pt, after=\vspace{-0.4cm},wide, labelwidth=!, labelindent=0pt]
\setlength\itemsep{-0.5em}
  \item Same
  \item Same
  \item Same
  \item Same
  \item the player 's agent said she had rejected the offer of a three-year contract
\end{enumerate}} \\ \hline
\small{\textit{Repeat (R-1=33.6,Rep=45):} \begin{enumerate}[topsep=-2pt,itemsep=-0.5pt, after=\vspace{-0.4cm},wide, labelwidth=!, labelindent=0pt]
\setlength\itemsep{-0.5em}
  \item Same
  \item Same
  \item Same
  \item Same
\end{enumerate}} & \small{\textit{Repeat (R-1=28, Rep=41.4):} 
\begin{enumerate}[topsep=-2pt,itemsep=-0.5pt, after=\vspace{-0.4cm},wide, labelwidth=!, labelindent=0pt]
\setlength\itemsep{-0.5em}
  \item Same
  \item Deleted
  \item Deleted
  \item Same
\end{enumerate}} & \small{\textit{Repeat (R-1=24.2, Rep=36.1):}
\begin{enumerate}[topsep=-2pt,itemsep=-0.5pt, after=\vspace{-0.4cm},wide, labelwidth=!, labelindent=0pt]
\setlength\itemsep{-0.5em}
  \item Same
  \item Same
  \item Deleted
  \item alves has been unable to agree a new deal
  \item Same
\end{enumerate}} \\ \hline
\small{\textit{Extra (R-1=43.6,Rep=36.8):}
\begin{enumerate}[topsep=-2pt,itemsep=-0.5pt, after=\vspace{-0.4cm},wide, labelwidth=!, labelindent=0pt]
\setlength\itemsep{-0.5em}
  \item Same
  \item Same
  \item the 31-year-old has been unable to agree a new deal with the catalan club and will leave the nou camp this summer
  \item Deleted
\end{enumerate}} & \small{\textit{Extra (R-1=37.2,Rep=92.8):}
\begin{enumerate}[topsep=-2pt,itemsep=-0.5pt, after=\vspace{-0.4cm},wide, labelwidth=!, labelindent=0pt]
\setlength\itemsep{-0.5em}
  \item Same
  \item Same
  \item Same
  \item Deleted
\end{enumerate}} & \small{\textit{Extra (R-1=37, Rep=60.8):} 
\begin{enumerate}[topsep=-2pt,itemsep=-0.5pt, after=\vspace{-0.4cm},wide, labelwidth=!, labelindent=0pt]
\setlength\itemsep{-0.5em}
  \item Same
  \item Same
  \item Same
  \item Same
  \item Deleted
\end{enumerate}} \\ \hline
\small{\textit{Mixture (R-1=43, Rep=36.2):} 
\begin{enumerate}[topsep=-2pt,itemsep=-0.5pt, after=\vspace{-0.4cm},wide, labelwidth=!, labelindent=0pt]
\setlength\itemsep{-0.5em}
  \item Same
  \item Same
  \item Same
  \item Deleted
\end{enumerate}} & \small{\textit{Mixture (R-1=28, Rep=41.43):}
\begin{enumerate}[topsep=-2pt,itemsep=-0.5pt, after=\vspace{-0.4cm},wide, labelwidth=!, labelindent=0pt]
\setlength\itemsep{-0.5em}
  \item Same
  \item Deleted
  \item Deleted
  \item Same
\end{enumerate}} & \small{\textit{Mixture (R-1=37, Rep=60.8):} 
\begin{enumerate}[topsep=-2pt,itemsep=-0.5pt, after=\vspace{-0.4cm},wide, labelwidth=!, labelindent=0pt]
\setlength\itemsep{-0.5em}
  \item Same
  \item Same
  \item Same
  \item Same
  \item Deleted
\end{enumerate}}
\\ \hline 
\end{tabularx}
\end{table*}
}

\clearpage

\section*{Acknowledgments}

We acknowledge support from the Swiss National Science Foundation (grant 31003A\_156976).


\begin{thebibliography}{}
\expandafter\ifx\csname natexlab\endcsname\relax\def\natexlab#1{#1}\fi

\bibitem[{Bahdanau et~al.(2014)Bahdanau, Cho, and Bengio}]{bahdanau2014neural}
Dzmitry Bahdanau, Kyunghyun Cho, and Yoshua Bengio. 2014.
\newblock Neural machine translation by jointly learning to align and
  translate.
\newblock {\em arXiv preprint arXiv:1409.0473\/} .

\bibitem[{Chatterjee et~al.(2018)Chatterjee, Negri, Rubino, and
  Turchi}]{chatterjee2018findings}
Rajen Chatterjee, Matteo Negri, Raphael Rubino, and Marco Turchi. 2018.
\newblock Findings of the wmt 2018 shared task on automatic post-editing.
\newblock In {\em Proceedings of the Third Conference on Machine Translation:
  Shared Task Papers\/}. pages 710--725.

\bibitem[{Chen and Bansal(2018)}]{chen2018fast}
Yen-Chun Chen and Mohit Bansal. 2018.
\newblock Fast abstractive summarization with reinforce-selected sentence
  rewriting.
\newblock In {\em Proceedings of ACL\/}.

\bibitem[{Erkan and Radev(2004)}]{erkan2004lexrank}
G{\"u}nes Erkan and Dragomir~R Radev. 2004.
\newblock Lexrank: Graph-based lexical centrality as salience in text
  summarization.
\newblock {\em Journal of artificial intelligence research\/} 22:457--479.

\bibitem[{Fevry and Phang(2018)}]{compression}
Thibault Fevry and Jason Phang. 2018.
\newblock \href{http://aclweb.org/anthology/K18-1040}{Unsupervised sentence
  compression using denoising auto-encoders}.
\newblock In {\em Proceedings of the 22nd Conference on Computational Natural
  Language Learning\/}. Association for Computational Linguistics, pages
  413--422.
\newblock
  \href{http://aclweb.org/anthology/K18-1040}{http://aclweb.org/anthology/K18-1040}.

\bibitem[{Gehrmann et~al.(2018)Gehrmann, Deng, and Rush}]{bottomup}
Sebastian Gehrmann, Yuntian Deng, and Alexander Rush. 2018.
\newblock \href{http://aclweb.org/anthology/D18-1443}{Bottom-up abstractive
  summarization}.
\newblock In {\em Proceedings of the 2018 Conference on Empirical Methods in
  Natural Language Processing\/}. Association for Computational Linguistics,
  pages 4098--4109.
\newblock
  \href{http://aclweb.org/anthology/D18-1443}{http://aclweb.org/anthology/D18-1443}.

\bibitem[{Hermann et~al.(2015)Hermann, Kocisky, Grefenstette, Espeholt, Kay,
  Suleyman, and Blunsom}]{hermann2015teaching}
Karl~Moritz Hermann, Tomas Kocisky, Edward Grefenstette, Lasse Espeholt, Will
  Kay, Mustafa Suleyman, and Phil Blunsom. 2015.
\newblock Teaching machines to read and comprehend.
\newblock In {\em Advances in Neural Information Processing Systems\/}. pages
  1693--1701.

\bibitem[{Isabelle(2007)}]{isabelle2007domain}
P~Isabelle. 2007.
\newblock Domain adaptation of mt systems through automatic postediting.
\newblock {\em Proc. 10th Machine Translation Summit (MT Summit XI), 2007\/} .

\bibitem[{Kingma and Ba(2015)}]{kingma2014adam}
Diederick~P Kingma and Jimmy Ba. 2015.
\newblock Adam: A method for stochastic optimization.
\newblock In {\em International Conference on Learning Representations
  (ICLR)\/}.

\bibitem[{Lin(2004)}]{lin2004rouge}
Chin-Yew Lin. 2004.
\newblock Rouge: A package for automatic evaluation of summaries.
\newblock {\em Text Summarization Branches Out\/} .

\bibitem[{Nallapati et~al.(2017)Nallapati, Zhai, and
  Zhou}]{nallapati2017summarunner}
Ramesh Nallapati, Feifei Zhai, and Bowen Zhou. 2017.
\newblock Summarunner: A recurrent neural network based sequence model for
  extractive summarization of documents.
\newblock In {\em Thirty-First AAAI Conference on Artificial Intelligence\/}.

\bibitem[{Narayan et~al.(2018)Narayan, Cohen, and Lapata}]{refresh}
Shashi Narayan, Shay~B. Cohen, and Mirella Lapata. 2018.
\newblock \href{https://doi.org/10.18653/v1/N18-1158}{Ranking sentences for
  extractive summarization with reinforcement learning}.
\newblock In {\em Proceedings of the 2018 Conference of the North American
  Chapter of the Association for Computational Linguistics: Human Language
  Technologies, Volume 1 (Long Papers)\/}. Association for Computational
  Linguistics, pages 1747--1759.
\newblock
  \href{https://doi.org/10.18653/v1/N18-1158}{https://doi.org/10.18653/v1/N18-1158}.

\bibitem[{Nenkova et~al.(2011)Nenkova, Maskey, and Liu}]{nenkova2011automatic}
Ani Nenkova, Sameer Maskey, and Yang Liu. 2011.
\newblock Automatic summarization.
\newblock In {\em Proceedings of the 49th Annual Meeting of the Association for
  Computational Linguistics: Tutorial Abstracts of ACL 2011\/}. Association for
  Computational Linguistics, page~3.

\bibitem[{Nikolov et~al.(2018)Nikolov, Pfeiffer, and Hahnloser}]{science-summ}
Nikola Nikolov, Michael Pfeiffer, and Richard Hahnloser. 2018.
\newblock Data-driven summarization of scientific articles.
\newblock In {\em Proceedings of the Eleventh International Conference on
  Language Resources and Evaluation (LREC 2018)\/}. European Language Resources
  Association (ELRA), Paris, France.

\bibitem[{Rozovskaya and Roth(2016)}]{rozovskaya2016grammatical}
Alla Rozovskaya and Dan Roth. 2016.
\newblock Grammatical error correction: Machine translation and classifiers.
\newblock In {\em Proceedings of the 54th Annual Meeting of the Association for
  Computational Linguistics (Volume 1: Long Papers)\/}. volume~1, pages
  2205--2215.

\bibitem[{See et~al.(2017)See, Liu, and Manning}]{see2017get}
Abigail See, Peter~J Liu, and Christopher~D Manning. 2017.
\newblock Get to the point: Summarization with pointer-generator networks.
\newblock In {\em Proceedings of the 55th Annual Meeting of the Association for
  Computational Linguistics (Volume 1: Long Papers)\/}. volume~1, pages
  1073--1083.

\bibitem[{Sennrich et~al.(2016{\natexlab{a}})Sennrich, Haddow, and
  Birch}]{sennrich2015improving}
Rico Sennrich, Barry Haddow, and Alexandra Birch. 2016{\natexlab{a}}.
\newblock \href{https://doi.org/10.18653/v1/P16-1009}{Improving neural machine
  translation models with monolingual data}.
\newblock In {\em Proc. of ACL\/}. Association for Computational Linguistics,
  pages 86--96.
\newblock
  \href{https://doi.org/10.18653/v1/P16-1009}{https://doi.org/10.18653/v1/P16-1009}.

\bibitem[{Sennrich et~al.(2016{\natexlab{b}})Sennrich, Haddow, and
  Birch}]{sennrich2015neural}
Rico Sennrich, Barry Haddow, and Alexandra Birch. 2016{\natexlab{b}}.
\newblock \href{https://doi.org/10.18653/v1/P16-1162}{Neural machine
  translation of rare words with subword units}.
\newblock In {\em Proc. of ACL\/}. Association for Computational Linguistics,
  pages 1715--1725.
\newblock
  \href{https://doi.org/10.18653/v1/P16-1162}{https://doi.org/10.18653/v1/P16-1162}.

\bibitem[{Sutskever et~al.(2014)Sutskever, Vinyals, and
  Le}]{sutskever2014sequence}
Ilya Sutskever, Oriol Vinyals, and Quoc~VV Le. 2014.
\newblock Sequence to sequence learning with neural networks.
\newblock In {\em Advances in neural information processing systems\/}. pages
  3104--3112.

\bibitem[{Xie et~al.(2018)Xie, Genthial, Xie, Ng, and Jurafsky}]{denoising}
Ziang Xie, Guillaume Genthial, Stanley Xie, Andrew Ng, and Dan Jurafsky. 2018.
\newblock \href{https://doi.org/10.18653/v1/N18-1057}{Noising and denoising
  natural language: Diverse backtranslation for grammar correction}.
\newblock In {\em Proceedings of the 2018 Conference of the North American
  Chapter of the Association for Computational Linguistics: Human Language
  Technologies, Volume 1 (Long Papers)\/}. Association for Computational
  Linguistics, pages 619--628.
\newblock
  \href{https://doi.org/10.18653/v1/N18-1057}{https://doi.org/10.18653/v1/N18-1057}.

\end{thebibliography}
\bibliographystyle{acl_natbib}

\end{document}